\documentclass[letterpaper]{article} 
\usepackage{aaai25}  
\usepackage{times}  
\usepackage{helvet}  
\usepackage{courier}  
\usepackage[hyphens]{url}  
\usepackage{graphicx} 
\usepackage{natbib}  
\usepackage{caption} 
\usepackage{algorithm}
\usepackage{algorithmic}

\usepackage{newfloat}
\usepackage{listings}
\DeclareCaptionStyle{ruled}{labelfont=normalfont,labelsep=colon,strut=off} 
\floatstyle{ruled}
\newfloat{listing}{tb}{lst}{}
\floatname{listing}{Listing}
\usepackage{soul}

\usepackage{xspace}

\def\our{\textsc{A-VL}\xspace}

\usepackage{booktabs}
\usepackage{multirow}

\usepackage{subcaption}

\title{\our: Adaptive Attention for Large Vision-Language Models}
\author {
    Junyang Zhang\textsuperscript{\rm 1},
    Mu Yuan\textsuperscript{\rm 1},
    Ruiguang Zhong\textsuperscript{\rm 2},
    Puhan Luo\textsuperscript{\rm 1},
    Huiyou Zhan\textsuperscript{\rm 1},
    Ningkang Zhang\textsuperscript{\rm 1},
    Chengchen Hu\textsuperscript{\rm 2},
    Xiang-Yang Li\textsuperscript{\rm 1}\thanks{Xiang-Yang Li is the corresponding author.}
}
\affiliations {
    \textsuperscript{\rm 1}University of Science and Technology of China, Hefei, China\\
    \textsuperscript{\rm 2}NIO Inc., Shanghai, China\\
    \{zhangjunyang, ym0813, luopuhan, zhangnk\}@mail.ustc.edu.cn, taurus.zhong@nio.com, zhanhuiyou@foxmail.com, huc@ieee.org, xiangyangli@ustc.edu.cn
}

\usepackage{bibentry}

\begin{document}

\maketitle

\begin{abstract}
The Large Vision-Language Model (LVLM) integrates computer vision and natural language processing techniques, offering substantial application potential. However, these models demand extensive resources during inference. 
Adaptive attention techniques can dynamically reduce computational redundancy and thus improve efficiency. 
Although current adaptive attention methods significantly reduce the memory requirements of Transformer-based language models, they are not tailored for LVLMs. 
We observe that LVLMs generate responses from both remote image tokens and local text tokens, and different modalities have different attention patterns. 
This observation inspires us to manage the attention for each modality separately. 
Specifically, for visual input, we store the cache of potentially useful information but only compute the most critical parts.
For language input, we care more about local information.
Based on our observation and analysis of vision-language attention patterns, we develop \our, a plug-and-play adaptive attention tailored for LVLM inference.
Extensive evaluations on three vision-language tasks and five datasets show the effectiveness of our designs.
Our approach \our outperforms existing adaptive attention methods in reducing memory usage and computational load without compromising performance.
\end{abstract}

%



\section{Introduction\label{sec:intro}}

In recent years, large vision-language models (LVLMs) have exhibited exceptional capabilities in vision-language understanding and reasoning~\cite{liu2024llavanext, Qwen-VL, wang2023cogvlm}. 
LVLMs are increasingly integrated into many applications, such as personal intelligent assistants and vehicle cockpit systems.
Despite their impressive performance, deploying LVLMs in real-world systems still faces practical challenges due to expensive resource overheads. 
Consequently, reducing computational overheads and improving inference speed have become critical problems.

In LVLM inference, visual and text inputs are separately encoded and fed into the pretrained large language model (LLM). 
As each token generated through LLM autoregression depends on all preceding tokens, this process consumes considerable time and memory resources.
This issue is exacerbated in LVLMs since high-resolution images lead to rapidly expanding token sequences.
Recent studies explored adaptive attention and KV cache compression techniques for single-modal language models, achieving significant advancements. 
For instance, StreamingLLM~\cite{xiao2023efficient} maintains only fixed cache positions, while H$_2$O~\cite{zhang2024h2o} preserves the most important caches based on the accumulated attention.
These methods reduce memory usage and enhance the inference speed of LLMs.
However, these methods are designed specifically for single-modal language models. 
Our experiments demonstrate the uniqueness of the attention pattern of LVLMs, which has inspired us to develop an adaptive attention scheme tailored to the unique needs of different modalities within LVLMs.

Adaptive attention methods must address two critical aspects: identifying redundancy and maintaining performance. 
To develop an effective adaptive attention method for LVLMs, we must first examine the specific characteristics of attention in LVLMs. 
Subsequently, we can design a strategy based on these characteristics to reduce memory usage and computational demands while maintaining performance.
In this paper, we reveal that different modalities of LVLMs exhibit different attention patterns, necessitating separate analysis and processing for each modality within LVLM's adaptive attention mechanism. 
Vision inputs, though positioned at the beginning of the sequence, consistently receive a substantial proportion of attention during generation. 
In contrast, text inputs at the sequence's end exhibit rapid attention decay.
Vision inputs also show patterns of attention sparsity and drift.
Specially, the attention at the token level for vision inputs presents a surprisingly sparse distribution. 
Additionally, we find that some useful vision caches are not required at every step. 
We observe a gradual shift in attention among these useful vision caches.

To address this issue, we develop distinct methods for different modalities. 
In response to the sparsity and attention drift observed in vision attention, we categorize attention importance and dynamically compute the most critical parts of the KV cache. 
Specifically, we first preserve caches likely to be useful in subsequent inference and evict those deemed useless. 
We then select the most critical caches in the current step from these useful caches for computation.
Simultaneously, we regularly update this critical cache to ensure accurate computations.
For text attention, which rapidly loses attention, we concentrate on the local cache and retain only the most essential remote text caches. 
Notably, our approach does not require fine-tuning the model and offers plug-and-play compatibility with LVLMs.

Our main contributions are listed as follows:
\begin{itemize}
    \item We investigate and identify different attention patterns across different modalities within LVLMs.
    \item We propose \our, a plug-and-play adaptive attention method for LVLMs, designed to significantly reduce memory usage and computational load during inference without compromising performance.
    \item We evaluate the effectiveness of \our across three representative vision-language tasks on different LVLMs.
    Experimental results show that \our outperforms existing adaptive attention methods in memory and computational efficiency.
\end{itemize}

\begin{figure*}[t]
\centerline{\includegraphics[width=1\linewidth]{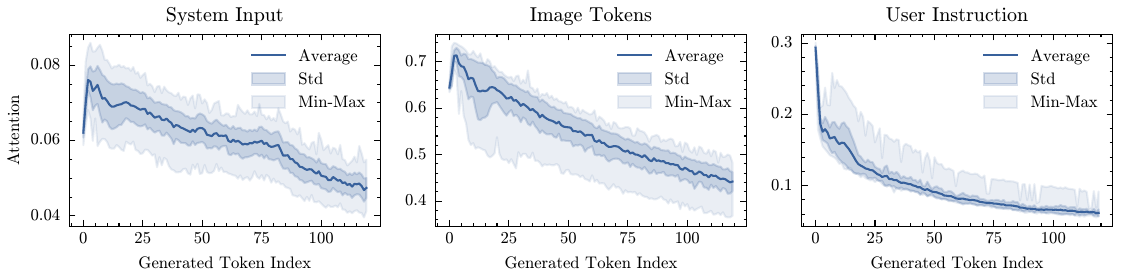}}
\caption{Variations in attention scores among different types of tokens during the decode phase.}
\label{fig:text_local}
\end{figure*}


\section{Background and Related Work\label{sec:background_and_related_work}}

\subsection{Large Vision-Language Modal.}

Large vision-language models (LVLMs) integrate the capabilities of visual encoders and large language models to understand multimodal information and generate textual responses related to images. 
LVLMs such as LLaVA~\cite{liu2024visual}, Qwen-VL~\cite{Qwen-VL} and CogVLM~\cite{wang2023cogvlm} have demonstrated impressive performance across various vision-language tasks. 
The LVLM typically consists of a visual encoder, an adapter, and a pre-trained LLM. 
The visual encoder processes and converts vision inputs into hidden features. 
Then the adapter transforms features into tokens for LLM.
Text inputs are also converted into tokens through the embedding model.
The LLM generates responses based on the combined visual and textual tokens. 
The inference of an LLM is divided into two phases: the prefill phase and the decode phase. 
During the prefill phase, all input tokens are processed and KV caches are simultaneously generated. 
The KV caches store the intermediate results of the processed tokens. 
In the decode phase, the model computes using the newly generated token and the KV cache from all previous tokens. 
As the sequence of tokens extends, the size of the KV cache grows linearly, directly affecting the memory usage and computational demands during inference.
However, the vision tokens from higher-resolution images output by the visual encoder significantly increases the length of token sequences.
For instance, LLaVA 1.5 processes 576 image tokens, while LLaVA 1.6 increases this number to over 2000. 
This increase presents challenges to the deployment and utilization of LVLMs.

\subsection{Inference Optimization for LLM and LVLM.}

The decoder-only transformer-based LLM necessitates autoregressive iteration for output generation, where the generation of each token depends on all preceding tokens. This dependency significantly increases memory requirements and computational overhead. To address these challenges, existing methods are categorized into two main approaches: inference process optimization and computation reduction. Inference process optimization includes strategies such as page attention~\cite{kwon2023efficient} and speculative sampling~\cite{xia2024unlocking} to achieve flexible and efficient management of the entire inference process. Computation reduction involves methods like adaptive attention, early exit~\cite{bae2023fast}, and structured pruning~\cite{an2024fluctuation}. 
Our primary focus is on adaptive attention technology, which dynamically reduces computational redundancy according to inputs and has produced remarkable results in LLMs.
StreamingLLM~\cite{xiao2023efficient} is designed to retain a fixed-position cache, which can train LLMs with long contexts on a limited cache window.
Anagnostidis et al.~\cite{anagnostidis2024dynamic} propose a new sigmoid function to dynamically sparse attention, but the model needs to be fine-tuned.
FastGen~\cite{ge2023model} combines multiple strategies to retain recent caches, special characters, and caches that have been historically assigned heavier attention.
H$_2$O~\cite{zhang2024h2o} leverages historical accumulated attention and combined with the local cache for KV cache reduction, and has achieved remarkable results.
Although these methods improve the inference efficiency of LLM, they are designed for single-modal text inputs.
In addition, there are some studies that focus on enhancing the efficiency of transformer-based visual models. SPViT~\cite{kong2022spvit} prunes tokens in Vision Transformers (ViT), and PuMer~\cite{cao2023pumer} prunes and merge tokens in cross attention; however, these methods are designed for encoder models. 
FastV~\cite{chen2024image} is an optimization solution designed for autoregressive LVLMs, revealing that without using KV cache, LVLM can potentially use only 50\% of tokens after the second layer. Nevertheless, the KV cache is an essential optimization tool for autoregressive inference. FastV has not been verified in inference with KV cache, because token pruning after the second layer may lead to the loss of some KV cache, making it impossible to select tokens in subsequent steps. And the pruning in FastV with the KV cache may lead to the loss of potentially important tokens in the subsequent steps.


\section{Observations and Insights}

In this section, we employ experiments to uncover key insights into the inference of large vision-language models, facilitating optimization of these processes.


\subsection{Attention of Different Modalities}

\begin{figure}[tb]
\centerline{\includegraphics[width=1\columnwidth]{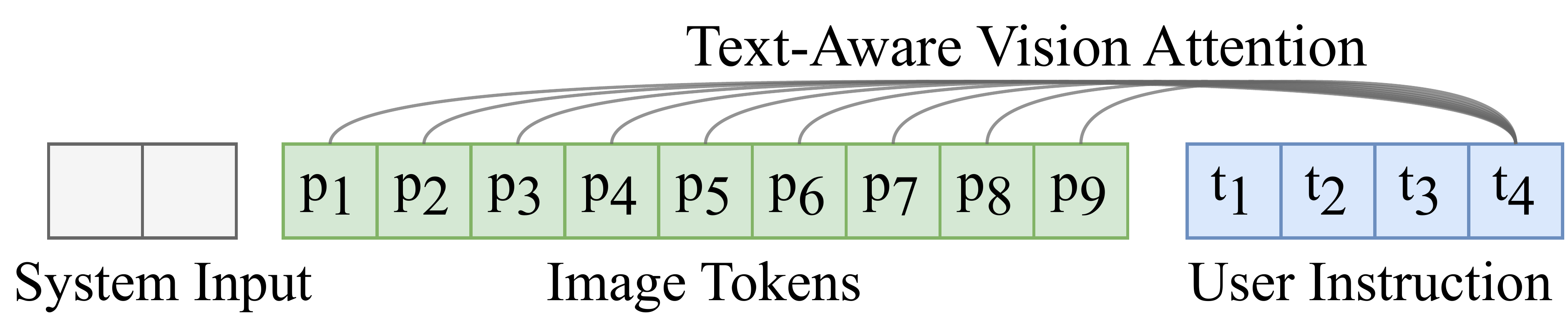}}
\caption{The token sequence is composed of encoded inputs from different modalities.}
\label{fig:token_demo}
\end{figure}

The large language model in LVLMs processes three types of input tokens: system input, image tokens and user instruction, as shown in Figure~\ref{fig:token_demo}. System input typically comprises control information, such as role settings and task modes. Image tokens represent vision information encoded by the visual encoder. User instruction is the query posed by users about the image. 
Attention in LLM quantifies the relevance between two tokens by assigning weights.
The attention directed toward image tokens in user queries implies the importance of these image tokens to the current question~\cite{cao2023pumer}, which we call text-aware vision attention.

We utilize the attention score to demonstrate the attention allocated to input tokens during the generation of a new token. The attention score is calculated as the average of the multi-head attention weights. The formula of attention score is as follows:
\begin{equation}
s^l_t = \sum^{H}_{h=1} A^{l,h}_{t} / H,
\label{eq:attention_score}
\end{equation}
where $A$ represents the attention weights assigned from the last token to all preceding tokens; $l$ denotes the index of the transformer decoder layers; and $h$ indicates the attention head.
To investigate attention changes in LVLM, we randomly select 100 samples from the Flickr30k dataset~\cite{young2014image} that can output longer text sequences. We quantify the attention score changes in LLaVA-1.5~7B model across three token types during the decode phase. The results are depicted in Figure~\ref{fig:text_local}.
Surprisingly, the attention allocated to system input and image tokens gradually decreases as the token sequence lengthens, while the attention to user instructions decreases sharply. 
Note that user instructions are often of similar length to system inputs or even longer, but still exhibit different attention patterns.
This suggests that the remote system inputs and image tokens are crucial for inference.
These results indicate that the attention allocated to each modality may need to be managed separately.

\subsection{Heterogeneity of Vision Attention}

\begin{figure}[tb]
\centerline{\includegraphics[width=1\columnwidth]{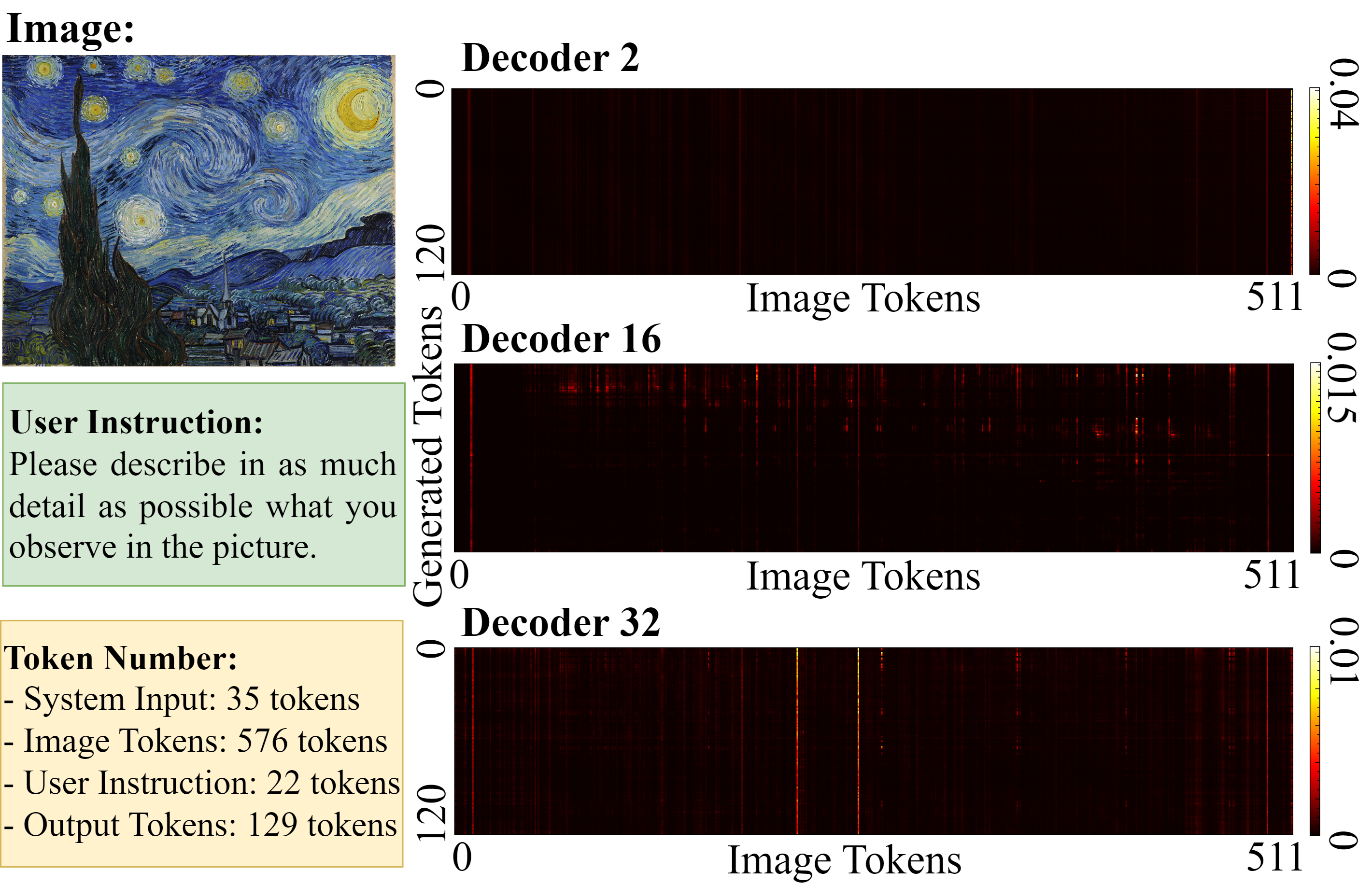}}
\caption{The vision attention during the generation of new tokens is very sparse.}
\label{fig:attention_demo}
\end{figure}

\begin{figure}[tb]
\centerline{\includegraphics[width=1\columnwidth]{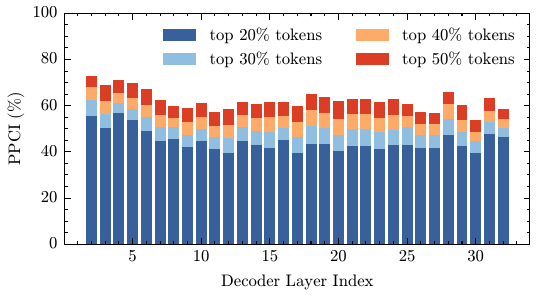}}
\caption{The correlation of attention between the first decoder layer and each subsequent decoder layer.}
\label{fig:attention_details}
\end{figure}

We subsequently explore how vision attention changes across different decoder layers.
For an intuitive visualization, we visualize the attention scores across various layers for image tokens during the inference with the LLaVA-1.5 7B model, as illustrated in Figure~\ref{fig:attention_demo}. The visualization shows that most image tokens are allocated little attention after first decoder and the vision attention is highly sparse.
To quantitatively illustrate the ubiquity of this sparsity, we use the LLaVA-1.5~7B model to evaluate the OCRVQA dataset and find that 22.88\% of the tokens receive over 80\% of the total attention in all layers.
This indicates that image information is highly concentrated in a few tokens, resulting in inefficient inference. It is similar to the Pareto principle. 
Based on the observation that a small number of image tokens are allocated a large amount of attention, we define a metric termed the $p$-percentile concordance index (PPCI).
This metric compares the correlation between two attention weights on the tokens, specifically focusing on key tokens. 
The PPCI measures the proportion of the top $p\%$ of tokens that are identical in two different attention weights.
Specifically, for two sequences of attention scores, $S_1$ and $S_2$, on the $N$ tokens, the sets $T_1$ and $T_2$ represent the sets of top $p\%$ tokens in $S_1$ and $S_2$ respectively, sorted by their attention scores. The $p$-percentile concordance index (PPCI) is defined as follows:
\begin{equation}
p\%\;PPCI = \frac{\mid T_1 \cap T_2 \mid}{p\% \cdot N} .
\label{eq:ppci}
\end{equation}
We use LLaVA-1.5 7B model to evaluate the OCRVQA dataset, and analyze the PPCI between the attention across each decoder layer and that in the first decoder layer.
The results are depicted in Figure~\ref{fig:attention_details}.
As illustrated by the figure, the 50\% PPCI between the first decoder layer and subsequent layers is less than 70\%, and the 50\% PPCI between the first layer and the last decoder layer is below 60\%.
We also calculate the 50\% PPCI between the first three layers and the last layer, finding it to be less than 70\% on average.
Based on this observation, we refer to this phenomenon where different decoder layers focus on different image tokens as the \textit{heterogeneity} of vision attention.
Consequently, it is essential to design the adaptive attention method based on the attention distribution for each layer.

\subsection{Continuity of Vision Attention}

\begin{figure}[tb]
\centerline{\includegraphics[width=1\columnwidth]{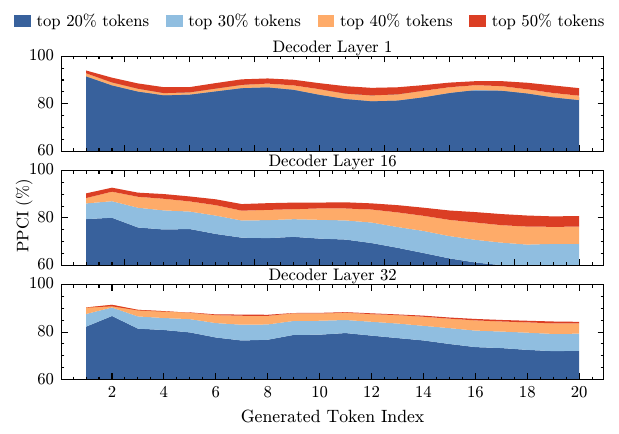}}
\caption{The correlation of attention between the first step and each subsequent step in the same decoder layer.}
\label{fig:continuity_attention}
\end{figure}

We subsequently analyze the attention from the temporal dimension with the LLaVA-1.5 7B model and the OCRVQA dataset. During the autoregressive generation of tokens, we observe a high correlation in attention within the same decoder layer across different steps, as depicted in Figure~\ref{fig:continuity_attention}.
The figure illustrates that during the generation of 20 new tokens (20 steps) in the same decoder layer, the $p\%$ PPCI in image tokens between each step and the first step.
As depicted in the figure, each layer consistently allocates high attention to the top 50\% of image tokens. For the top 30\% image tokens, each layer allocates high attention in the short term, although this may shift after three steps. 
We refer to this short-term correlation of attention among image tokens within the same layer as the \textit{continuity} of vision attention.
This indicates that attention may gradually drift within partial image tokens. 
Given the gradual nature of these changes, we can reasonably assume that the set of core tokens remains stable in the short term.

Based on the heterogeneity and continuity of vision attention, we can select important image tokens for each layer according to the attention scores, thereby reducing computational load.
In fact, we retain only the top 30\% of image tokens for each decoder layer using the LLaVA-1.5 7B model and OCRVQA dataset to ensure that 99\% of the samples generate three tokens identical to the original model output.
However, for longer outputs, because attention drifts, relying on only 30\% image tokens is not enough to predict.
Thus, we can preserve potentially useful image tokens for future use, but compute only the most critical token at each step. 
This specific design will be elaborated in the following section.

\subsection{Locality of Text Attention}

As illustrated in Figure~\ref{fig:text_local}, the attention of text tokens decays rapidly. 
Therefore, it is advisable to focus on local text attention and a limited number of critical remote text attention.
The H$_2$O method can be precisely applied to achieve this objective. Specially, we maintain local text cache caches and evict useless text caches based on accumulated historical attention.
We also analyze the attention patterns of Llama 2~\cite{touvron2023llama}, a large language model. 
We observe that the attention patterns in Llama 2 are similar to those in the text token of LVLMs, both characterized by rapid attention decay.
This similarity further underscores the suitability of employing an LLM's adaptive attention method to the text tokens of LVLMs.


\section{\our Design \label{sec:system_design}}

Based on the above insights, we propose \our, a plug-and-play adaptive attention method tailored for LVLMs.

\begin{figure}[tb]
\centerline{\includegraphics[width=1\linewidth]{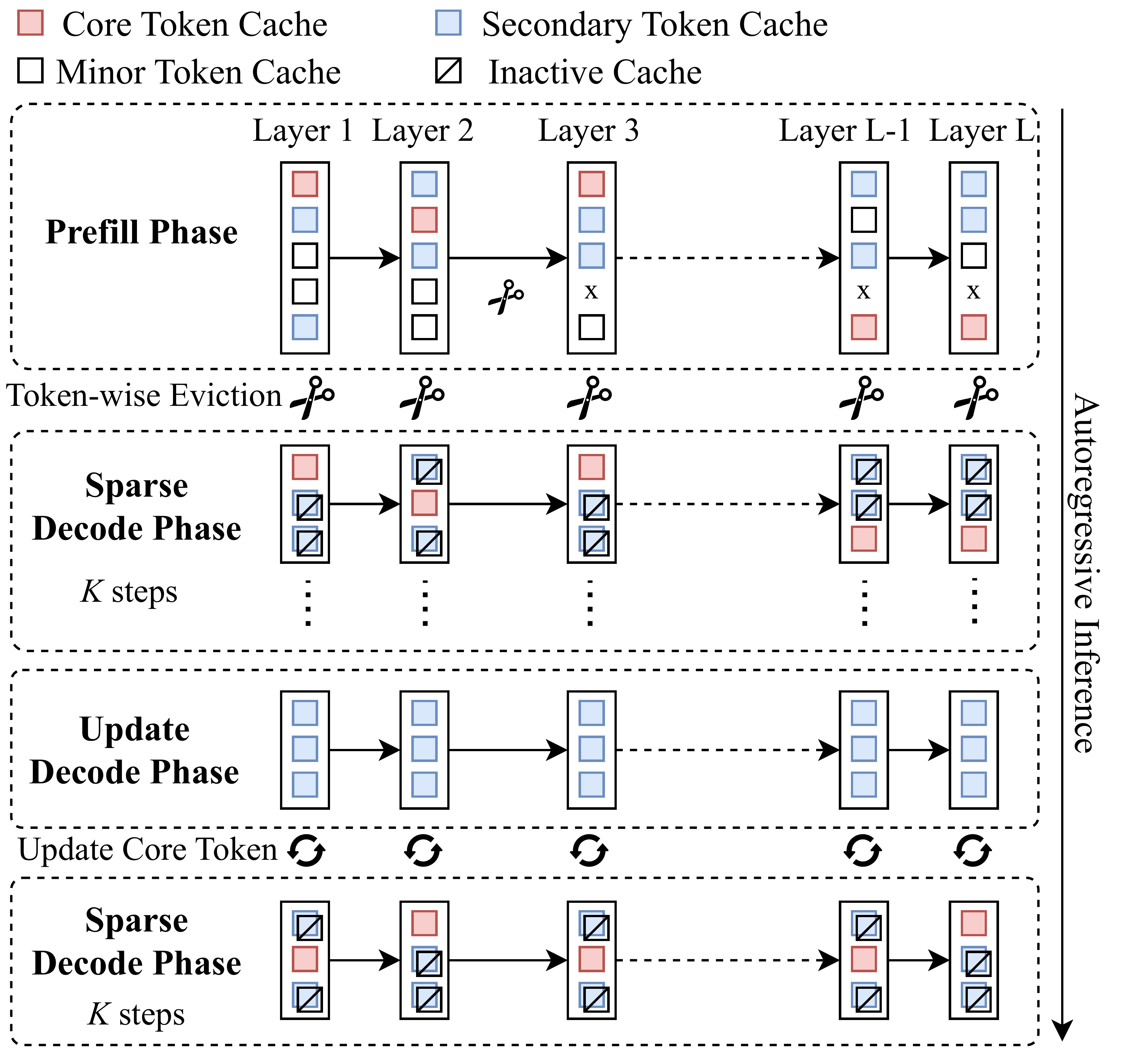}}
\caption{The design of adaptive vision attention.}
\label{fig:image_pruning}
\end{figure}

\subsection{Adaptive Text-Aware Vision Attention}

\subsubsection{Decode phase.}
We design a hierarchical adaptive attention mechanism for vision attention based on the attention scores from the perspective of text tokens, as shown in Figure~\ref{fig:image_pruning}.
In the prefill phase, we categorize image tokens into three sets: core, secondary, and minor, according to the attention score of the last text token.
This classification uses the attention score from the text token to emphasize the semantic relationship between text and images.
Specifically, after computing the attention score in each decoder layer, tokens within the top $C\%$ of scores are designated as core tokens, those within the top $S\%$ as secondary tokens, and the remainder as minor tokens. 
We ensure that $S > C$, thereby making core tokens a subset of secondary tokens.
Minor tokens are immediately evicted to minimize KV cache usage. Secondary tokens, which may be useful in the future, are retained in memory; however, they are not all used in computations. 
Only core tokens are involved in the computations of subsequent attention modules.
However, our observations suggest that vision attention may shift gradually among secondary tokens, potentially altering the set of core tokens. 
Thus, every $K$ steps, we utilize all secondary tokens for inference and compute the attention scores to update the core token set, ensuring that attention remains focused on the most significant tokens. We refer to this step as the update decode phase.
This design is based on our insights into the continuity of vision attention, allowing the model to minimize cache usage and further reduce computational load of attention.
It should be noted that our method is applied independently to each decoder layer, and the results are not utilized until running to the corresponding layer in the next step, allowing for parallel processing with the original model's inference.

\subsubsection{Prefill phase.}
The related work, FastV, has shown that inference without using KV cache may result in the deletion of half image tokens based on attention scores at each step.
However, employing KV cache with this method leads to performance degradation, as potentially useful tokens are evicted, causing cache misses.
Our experiments illustrate this phenomenon in evaluation section.
Inspired by this study, we propose simplifying this technology for only use in the prefill phase.
Specifically, we only retain $P\%$ image tokens based on attention score after the second decoder layer in the prefill phase. And then we use only the KV cache of these retained image tokens for inference in the decode phase.
We can reduce the parameter $P$ to prevent the performance from decreasing.
This technology reduces computational load in the prefill stage and has been experimentally verified to be compatible with the above adaptive vision attention technology.

\subsection{Adaptive Text Attention}

Based on the insight of text attention, when generating new text tokens, both remote image tokens and local text tokens are used to predict. 
Since image tokens are processed separately, we use H$_2$O's accumulated attention method for text cache eviction.
Specially, we set a text cache window size of $T\%$ of the maximum text token length. 
After computing attention weights at each layer, we separate the attention of vision and text, and store the accumulated text attention scores.
If the text cache size exceeds the set text cache window size, we evict the redundant caches from the first half of the window based on these scores.
This method, utilized by H$_2$O, has proven effective in LLM.
We incorporate this approach into the text attention of LVLMs and verify its effectiveness through experiments.
We list all the parameters involved in our method and their description in Table~\ref{tab:hyperparameters}.

\begin{figure}[tb]
\centerline{\includegraphics[width=1\columnwidth]{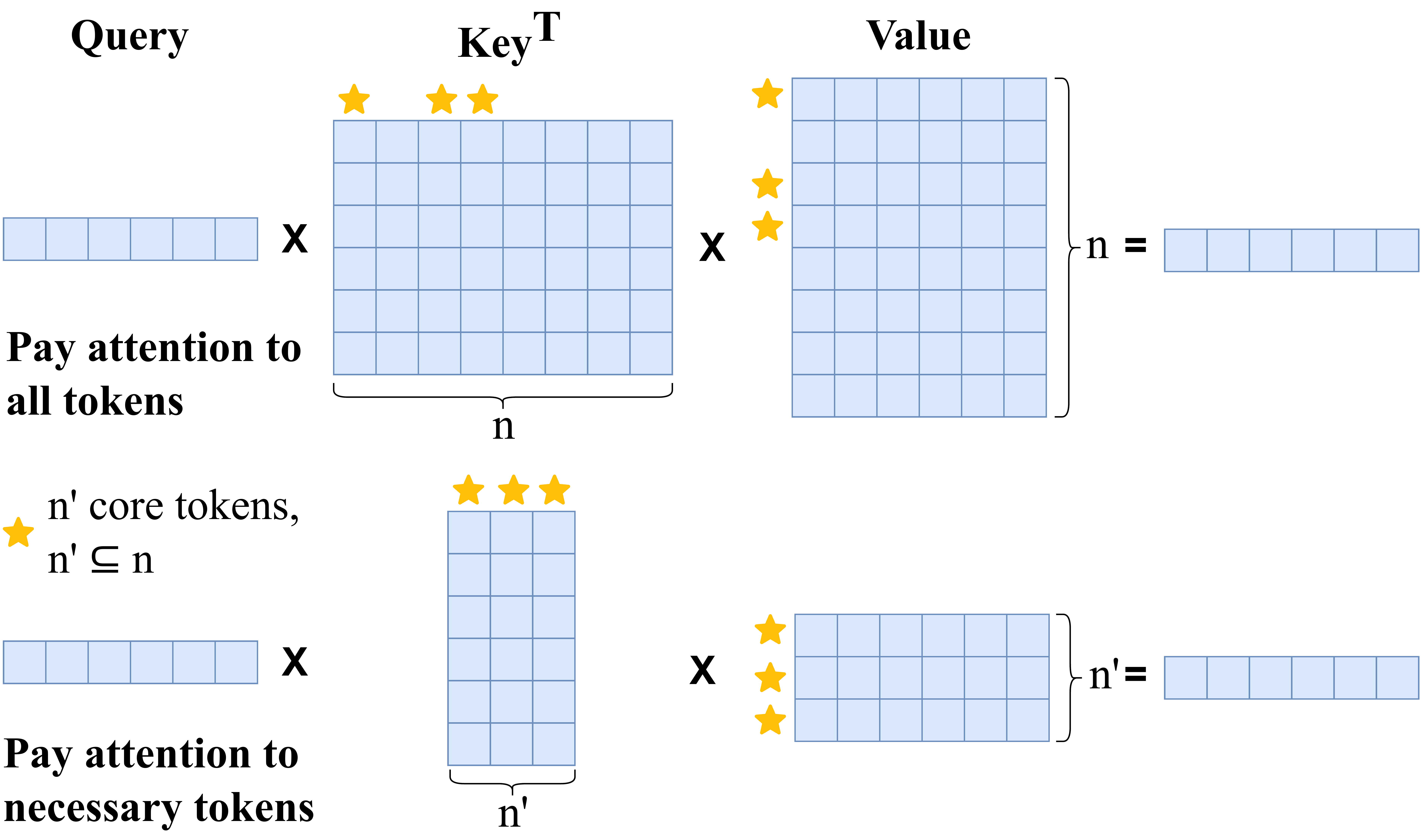}}
\caption{Adaptive vision attention uses only core KV cache. The details such as softmax are omitted in the figure.}
\label{fig:adaptive_attention}
\end{figure}

\begin{table}[b]
\centering
\begin{tabular}{@{}cl@{}}
\toprule
\textbf{Param.} & \textbf{Description}                                       \\ \midrule
S\%               & The proportion of secondary image tokens                   \\
C\%               & The proportion of core image tokens                        \\
K                 & Update the core token set every K steps                    \\
P\%               & Image tokens retained in prefill phase \\
T\%               & The proportion of retained text cache                    \\ \bottomrule
\end{tabular}
\caption{Description of parameters in \our.}
\label{tab:hyperparameters}
\end{table}

\begin{table*}[ht]
\centering
\begin{tabular}{@{}c|c|ccccccc@{}}
\toprule
\textbf{Model} & \textbf{Method} & \textbf{DocVQA} & \textbf{TextVQA} & \textbf{OCRBench} & \textbf{Nocaps} & \textbf{Flickr30k} & \textbf{VQAv2} & \textbf{Avg.} \\ \midrule
\multirow{4}{*}{LLaVA 1.5 7B}  & Original     & 31.20 & 48.67 & 20.60 & 101.88 & 73.81 & 76.52 & \textbf{58.78} \\
                               & FastV        & 27.35 & 47.16 & 19.30 & 101.06 & 72.99 & 74.86 & 57.12 \\
                               & H2O          & 30.34 & 48.22 & 20.40 & 102.23 & 73.53 & 76.49 & 58.54 \\
                               & \textbf{Our} & 31.26 & 48.44 & 20.90 & 102.67 & 73.81 & 76.52 & \textbf{58.93} \\ \midrule
\multirow{4}{*}{LLaVA 1.5 13B} & Original & 35.39 & 52.86 & 22.50 & 107.53 & 79.69 & 77.97 & \textbf{62.66} \\
                               & FastV        & 30.27 & 50.64 & 20.60 & 106.41 & 78.88 & 76.65 & 60.57 \\
                               & H2O          & 33.86 & 52.72 & 22.30 & 107.11 & 79.15 & 77.96 & 62.18 \\
                               & \textbf{Our} & 35.22 & 52.97 & 22.40 & 107.52 & 79.79 & 77.97 & \textbf{62.65} \\ \midrule
\multirow{4}{*}{LLaVA 1.6 7B}  & Original & 74.43 & 64.65 & 53.10 & 88.18  & 68.45 & 80.01 & \textbf{71.47} \\
                               & FastV        & 67.69 & 63.15 & 48.30 & 86.21  & 66.73 & 79.58 & 68.61 \\
                               & H2O          & 72.01 & 64.15 & 51.10 & 88.63  & 68.61 & 79.97 & 70.75 \\
                               & \textbf{Our} & 74.46 & 64.69 & 52.00 & 88.14  & 68.38 & 80.02 & \textbf{71.28} \\ \midrule
\multirow{4}{*}{LLaVA 1.6 13B} & Original & 77.23 & 67.06 & 53.90 & 88.12  & 66.67 & 80.93 & \textbf{72.32} \\
                               & FastV        & 70.26 & 65.14 & 49.10 & 87.78  & 65.96 & 80.48 & 69.79 \\
                               & H2O          & 75.85 & 66.58 & 53.00 & 88.33  & 66.71 & 80.91 & 71.90 \\
                               & \textbf{Our} & 77.30 & 66.97 & 53.90 & 87.68  & 66.52 & 80.94 & \textbf{72.22} \\ \midrule
\multirow{4}{*}{Qwen-VL} & Original           & 68.65 & 61.10 & 49.10 & 58.92  & 59.01 & 79.17 & \textbf{62.66} \\
                               & FastV        & 51.45 & 52.85 & 38.60 & 58.73  & 57.84 & 76.05 & 55.92 \\
                               & H2O          & 65.84 & 60.59 & 46.60 & 59.06  & 59.39 & 79.16 & 61.77 \\
                               & \textbf{Our} & 68.09 & 60.66 & 49.10 & 58.46  & 58.88 & 79.16 & \textbf{62.39} \\
                               \bottomrule
\end{tabular}
\caption{Performance comparison of different models, different tasks and different methods. Except for the original model, each method only retains 50\% of the KV cache. The first and second highest average metrics are bolded.}
\label{tab:performance}
\end{table*}

\subsection{Implementation of \our}

\begin{figure}[tb]
\centerline{\includegraphics[width=1\columnwidth]{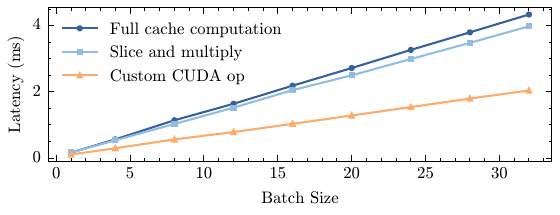}}
\caption{The latency to calculate 30\% caches in attention.}
\label{fig:multiply_latency}
\end{figure}

Our method is plug-and-play, requiring no fine-tuning of the original model. 
Cache selection and eviction can be processed in parallel with the original model inference, and the results are not utilized until the next step.
Besides, in our adaptive vision attention method, core caches are selected from secondary caches for computation, as depicted in Figure~\ref{fig:adaptive_attention}.
Slicing the cache before matrix multiplication introduces a performance bottleneck, because the core caches are not stored contiguously in memory~\cite{liu2023deja}.
Sometimes increasing latency of slicing beyond original matrix multiplication.
To address this, we develop a specialized CUDA operator that allows direct multiplication with selected rows or columns of the second matrix, eliminating the need for slicing. 
This CUDA operator employs block processing and shared memory to enhance speed further.
We measure the latency during the process in Figure~\ref{fig:adaptive_attention} with PyTorch on an NVIDIA A40 GPU under the configure of LLaVA-1.6 7B. 
Latency variations across different batch sizes are illustrated in Figure~\ref{fig:multiply_latency}.
It is evident that slicing is slow and negates the benefits of reduced computation. Our CUDA operator demonstrates effective acceleration.

\section{Evaluation\label{sec:evaluation}}

In this section, we demonstrate the effectiveness of our method through experiments.

\begin{table*}[ht]
\setlength{\tabcolsep}{1.85mm}
\centering
\begin{tabular}{c|ccccc|ccc|cccc}
\hline
\multirow{2}{*}{\textbf{Model}} & \multicolumn{5}{c|}{\textbf{Settings}} & \multirow{2}{*}{\textbf{\begin{tabular}[c]{@{}c@{}}Stored\\ Cache\end{tabular}}} & \multirow{2}{*}{\textbf{\begin{tabular}[c]{@{}c@{}}Used\\ Cache\end{tabular}}} & \multirow{2}{*}{\textbf{\begin{tabular}[c]{@{}c@{}}KV Cache\\ Memory\end{tabular}}} & \multirow{2}{*}{\textbf{DocVQA}} & \multirow{2}{*}{\textbf{TextVQA}} & \multirow{2}{*}{\textbf{OCRBench}} & \multirow{2}{*}{\textbf{Nocaps}} \\ \cline{2-6}
 & \textbf{S\%} & \textbf{C\%} & \textbf{K} & \textbf{P\%} & \textbf{T\%} &  &  &  &  &  &  &  \\ \hline
\multirow{5}{*}{\begin{tabular}[c]{@{}c@{}}LLaVA\\ 1.5 7B\end{tabular}} & - & - & - & - & - & 100\% & 100\% & 339 MB & 31.20 & 48.67 & 20.60 & 101.88 \\
 & 40 & - & - & - & - & 46\% & 46\% & 156 MB & 31.26 & 48.44 & 20.90 & 102.67 \\
 & 40 & 30 & 3 & - & - & 46\% & 39\% & 156 MB & 31.02 & 48.44 & 21.00 & 102.67 \\
 & 45 & 30 & 3 & 90 & - & 50\% & 40\% & 170 MB & 31.02 & 48.46 & 20.70 & 102.34 \\
 & 45 & 30 & 3 & 90 & 50 & 48\% & 38\% & 163 MB & 30.92 & 48.46 & 20.50 & 102.22 \\ \hline
\multirow{5}{*}{\begin{tabular}[c]{@{}c@{}}LLaVA\\ 1.5 13B\end{tabular}} & - & - & - & - & - & 100\% & 100\% & 529 MB & 35.39 & 52.86 & 22.50 & 107.53 \\
 & 40 & - & - & - & - & 46\% & 46\% & 243 MB & 35.37 & 52.86 & 22.20 & 107.61 \\
 & 45 & 30 & 3 & - & - & 50\% & 40\% & 265 MB & 35.22 & 52.97 & 22.40 & 107.52 \\
 & 45 & 30 & 3 & 90 & - & 50\% & 40\% & 265 MB & 35.46 & 52.83 & 22.30 & 107.60 \\
 & 45 & 30 & 3 & 90 & 60 & 48\% & 38\% & 254 MB & 35.43 & 52.84 & 22.30 & 107.59 \\ \hline
\multirow{4}{*}{\begin{tabular}[c]{@{}c@{}}LLaVA\\ 1.6 7B\end{tabular}} & - & - & - & - & - & 100\% & 100\% & 1179 MB & 74.43 & 64.65 & 53.10 & 88.18 \\
 & 40 & - & - & - & - & 43\% & 43\% & 507 MB & 74.29 & 64.76 & 51.20 & 88.14 \\
 & 45 & 30 & 3 & 90 & - & 47\% & 36\% & 554 MB & 74.46 & 64.71 & 51.90 & 87.43 \\
 & 45 & 30 & 3 & 90 & 70 & 45\% & 35\% & 531 MB & 74.44 & 64.69 & 52.00 & 87.45 \\ \hline
\multirow{4}{*}{\begin{tabular}[c]{@{}c@{}}LLaVA\\ 1.6 13B\end{tabular}} & - & - & - & - & - & 100\% & 100\% & 1841 MB & 77.23 & 67.06 & 53.90 & 88.12 \\
 & 40 & - & - & - & - & 43\% & 43\% & 792 MB & 77.30 & 66.97 & 53.90 & 87.68 \\
 & 45 & 30 & 3 & 90 & - & 47\% & 36\% & 865 MB & 76.97 & 67.06 & 54.40 & 87.59 \\
 & 45 & 30 & 3 & 90 & 70 & 45\% & 35\% & 828 MB & 76.97 & 67.06 & 54.50 & 87.59 \\ \hline
\end{tabular}
\caption{Performance under different settings with our method. KV Cache Memory refers to the memory usage when the batch size is one. As the batch size increases, memory consumption scales proportionally, resulting in greater memory savings.}
\label{tab:settings}
\end{table*}


\subsection{Evaluation Setup}

We implement our evaluation system based on the multimodal evaluation suite, LMMs-Eval~\cite{zhang2024lmms}.

\subsubsection{Evaluation Tasks} We employ multiple vision-language tasks to evaluate our method. Since our method primarily optimizes the KV cache in the decode phase, we do not choose any multiple-choice datasets that generate only one token during the prefill phase. For example, SEED-Bench~\cite{li2024seed} is a fully multiple-choice benchmark, so we did not select it. The tasks and datasets used in our evaluation are as follows:
\begin{itemize}
    \item Image Caption. This task involves automatically generating textual descriptions for visual content. We employ two datasets, Nocaps~\cite{agrawal2019nocaps} and Flickr30k~\cite{plummer2015flickr30k}. The metric is the CIDEr score~\cite{vedantam2015cider}.
    \item Visual Question Answering (VQA). This task requires models to answer questions based on visual information from images. We use three datasets, DocVQA~\cite{mathew2021docvqa}, TextVQA~\cite{singh2019towards} and VQAv2~\cite{DBLP:conf/cvpr/GoyalKSBP17}, measuring performance with the ANLS~\cite{9011031} and accuracy metrics.
    \item Optical Character Recognition (OCR). OCR involves recognizing text content within images. We select the OCRBench dataset~\cite{liu2023hidden}, which is specifically designed for LVLMs. The metric is the score designed in OCRBench.
\end{itemize}

Among these tasks, some emphasize the overall content of the image (e.g., image caption), and some focus on fine-grained details (e.g., OCR). This distinction enables the evaluation across varying levels of granularity.

\subsubsection{Models.} We use mainstream LVLMs and their different sizes, including LLaVA 1.5 7B, LLaVA 1.5 13B, LLaVA 1.6 7B, LLaVA 1.6 13B and Qwen-VL.

\subsubsection{Baselines.} We compare the output of the original model, FastV designed for LVLMs, and H$_2$O designed for LLMs. For the H$_2$O method, set the cache window to 75\% recent and 25\% heavy hitter. For FastV, in order to compare performance with the KV cache, we delete redundant tokens according to the FastV method in the prefill phase, and use the generated KV cache for inference in the decode phase.

\begin{figure}[tb]
    \centering
    \begin{subfigure}[b]{0.48\linewidth}
        \centering
        \includegraphics[width=\linewidth]{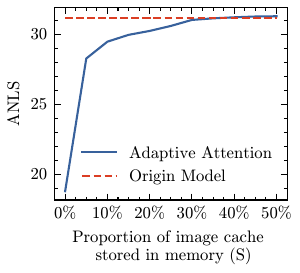}
        \caption{different secondary cache.}
        \label{fig:para_S}
    \end{subfigure}
    \hfill
    \begin{subfigure}[b]{0.50\linewidth}
        \centering
        \includegraphics[width=\linewidth]{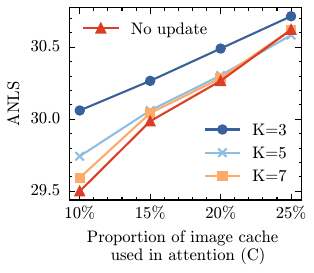}
        \caption{different core cache.}
        \label{fig:para_C}
    \end{subfigure}
    \caption{Performance under different parameters S, C, K.}
    \label{fig:para_CS}
\end{figure}

\subsection{Performance Guarantee}

We evaluate the performance of various models across different tasks, as shown in Table~\ref{tab:performance}.
FastV, H$_2$O, and our method retains 50\% of the KV cache. 
Our method achieves nearly lossless performance using only 50\% of the KV cache.
And our average performance is better than the other two baselines.
FastV shows reduced performance when utilizing the KV cache, due to the loss of potentially useful tokens during token pruning.
The H$_2$O method designed for LLMs shows effectiveness, but its average performance is still inferior to our method.
We observe that while H$_2$O can sometimes maintain performance like our method in image caption tasks that focus on the overall image information, our method demonstrates better performance in OCR and VQA tasks, which require attention to finer details. This indicates that reducing the number of image tokens may have minimal impact on tasks emphasizing general image content, but for detail-oriented tasks, our method, specifically designed for LVLMs is essential.
Our method demonstrates good results across various models, confirming its effectiveness.
In all methods, 50\% of the KV Cache is retained, resulting in similar decoding inference latency. Detailed memory savings from the KV Cache are shown in Table~\ref{tab:settings}.

\begin{figure}[tb]
\centerline{\includegraphics[width=1\columnwidth]{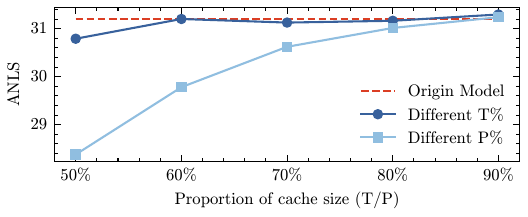}}
\caption{Performance under different parameters P and T.}
\label{fig:para_PT}
\end{figure}

\subsection{Memory Usage of KV Cache}

We first analyze the relationship between KV cache memory usage and performance across various parameters with LLaVA-1.5~7B model and DocVQA dataset. 
For vision attention, we evaluate the performance with different parameters, as shown in Figure~\ref{fig:para_CS}. 
Figure~\ref{fig:para_S} illustrates that retaining 40\% of the image cache as the secondary cache guarantees no performance loss, and retaining only 20\% results can ensure performance decline within 3\%. 
According to Figure~\ref{fig:para_C}, computing with only the core cache yields better results than simply eliminating these caches. 
Additionally, increasing the update frequency (lowering K) ensures that the core cache consistently contains the most critical elements.
These results confirm the insights previously discussed.
Performance under varying parameters P and T are presented in Figure~\ref{fig:para_PT}. 
Specifically, maintaining over 60\% of the text cache (Parameter T) ensures stable performance.
In contrast, retaining less than 80\% of the image tokens during the prefill phase (Parameter P) leads to performance degradation.
This underscores the underperformance of the FastV method with the KV cache, as previously analyzed.
However, integration with our method can reduce computational load by nearly 10\% during the prefill phase.

Our comprehensive analysis of different models and datasets under various settings is shown in Table~\ref{tab:settings}. 
As computational and memory demands decrease, the performance decreases slightly.
Nevertheless, our method maintains near-lossless performance with less than 50\% cache stored and only 35\% used in computations.
Notably, our method does not utilize the entire stored cache for computations, thus reducing the computational load compared to other methods, offering a unique advantage not provided by the baselines.

\subsection{Decoder Inference Latency}

Our method utilizes fixed eviction ratios for each layer, facilitating batch inference. We measure the inference latency of the Transformer Decoder in LLaVA-1.6 7B under various batch sizes on NVIDIA A40.  
By compressing the KV cache to 50\%, we achieve a 1.8x increase in decoder inference. Furthermore, by utilizing our CUDA operator to only calculate the core cache, we can further enhance the speed by an additional 1.1x. Consequently, the total decoder latency required is only 50.5\% of the original.
It is evident that our method can substantially reduce inference latency.
While our custom CUDA operator does increase speed, the improvement is less pronounced than in isolated tests due to the influence of other operations in PyTorch on its performance. We plan to further improve our CUDA operator in future work.


\section{Conclusion\label{sec:conclusion}}

In this paper, we propose \our, a plug-and-play adaptive attention method tailored specifically for large vision-language models. 
Our analysis of attention patterns across different modalities indicates that remote vision tokens consistently receive high attention, while textual tokens exhibit rapid attention decay.
Moreover, vision attention patterns are characterized by sparsity and drift, enabling dynamic selection of the most critical cache for computation. 
Consequently, we design the adaptive attention method tailored to each modality separately, which reduces the KV cache stored and concentrates computational load on the most critical cache. 
Our experiments confirm the efficacy of \our, demonstrating substantial reductions in memory usage and computation load without compromising performance.


\section{Acknowledgments}
The research is partially supported by USTC-NIO Smart Electric Vehicle Joint Lab, National Key R\&D Program of China under Grant No. 2021ZD0110400, Innovation Program for Quantum Science and Technology 2021ZD0302900 and China National Natural Science Foundation with No. 62132018, 62231015, 623B2093, "Pioneer" and "Leading Goose" R\&D Program of Zhejiang, 2023C01029 and 2023C01143.

\bibliography{aaai25}

\end{document}